\documentclass[10pt,twocolumn,letterpaper]{article}

\usepackage[pagenumbers]{wacv} 

\usepackage{graphicx}
\usepackage{amsmath}
\usepackage{amssymb}
\usepackage{booktabs}

\usepackage{multirow}
\usepackage{enumitem}
\usepackage{verbatim}
\usepackage{comment}
\usepackage{bm}
\usepackage{adjustbox}
\usepackage{array}
\usepackage{pifont}
\usepackage{stmaryrd}
\usepackage{booktabs}
\usepackage{flushend}
\usepackage{makecell}
\usepackage{float}
\usepackage{xfrac}
\usepackage{rotating}
\usepackage{colortbl}
\usepackage{xspace}

\usepackage{pifont}
\usepackage{setspace}
\usepackage{cuted}
\usepackage{bm}
\usepackage{diagbox}
\usepackage{svg}
\usepackage{balance}

\usepackage{amsfonts}  
\usepackage{nicefrac}       %
\usepackage{microtype} 
\usepackage[normalem]{ulem} %
\usepackage{xcolor}

\usepackage[linesnumbered,ruled,vlined]{algorithm2e}
\usepackage{algpseudocode}
\SetKwInput{KwInput}{Input}

\newcommand{\vz}{\mathbf{z}}
\newcommand{\vZ}{\mathbf{Z}}
\newcommand{\vt}{\mathbf{t}}

\newcommand{\vx}{\mathbf{x}}

\newcommand{\vell}{\boldsymbol{\ell}}

\newcommand{\img}{\scriptscriptstyle img}
\newcommand{\txt}{\scriptscriptstyle txt}

\definecolor{mygray}{gray}{0.5}

\SetCommentSty{mycommfont}

\DeclareMathOperator*{\argmax}{arg\,max}

\newcommand\gianni{\textcolor{black}}
\newcommand\clement{\textcolor{black}}

\newcommand{\andrei}[1]{\textcolor{black}{#1}}
\newcommand{\crot}{\rotatebox[origin=c]{90}}

\usepackage[capitalize]{cleveref}
\crefname{section}{Sec.}{Secs.}
\Crefname{section}{Section}{Sections}
\Crefname{table}{Table}{Tables}
\crefname{table}{Tab.}{Tabs.}

\def\wacvPaperID{1317} %
\def\confName{WACV}
\def\confYear{2024}

\begin{document}

\title{Improving CLIP Robustness with Knowledge Distillation and Self-Training}

\author{Clement Laroudie\\
U2IS, ENSTA Paris, Institut polytechnique de Paris\\
{\tt\small clement.laroudie@ensta-paris.fr}
\and 
Andrei Bursuc\\
valeo.ai\\
{\tt\small andrei.bursuc@valeo.com}
\and
Mai Lan Ha\\
Bosch Center for Artificial Intelligence\\
{\tt\small mailan.ha@de.bosch.com}
\and
Gianni Franchi\\
U2IS, ENSTA Paris, Institut polytechnique de Paris\\
{\tt\small gianni.franchi@ensta-paris.fr}
}


\maketitle

\begin{abstract}
\andrei{This paper examines the robustness of a multi-modal computer vision model, CLIP (Contrastive Language-Image Pretraining), in the context of unsupervised learning. 
The main objective is twofold: first, to evaluate the robustness of CLIP, and second, to explore strategies for augmenting its robustness. 
To achieve this, we introduce a novel approach named LP-CLIP. This technique involves the distillation of CLIP features through the incorporation of a linear probing layer positioned atop its encoding structure. This newly added layer is trained utilizing pseudo-labels produced by CLIP, coupled with a self-training strategy.
The LP-CLIP technique offers a promising approach to enhance the robustness of CLIP without the need for annotations. By leveraging a simple linear probing layer, we aim to improve the model's ability to withstand various uncertainties and challenges commonly encountered in real-world scenarios. Importantly, our approach does not rely on annotated data, which makes it particularly valuable in situations where labeled data might be scarce or costly to obtain. Our  proposed approach increases the robustness of CLIP with SOTA results compared to supervised technique on various datasets.}

\end{abstract}

\section{Introduction}

Foundation models  have emerged as powerful tools for processing and understanding various data modalities, encompassing text and images. These models, which include GPT-3 \cite{brown2020language}, LLaMa \cite{touvron2023llama}, and CLIP \cite{radford2021learning}, among others, are Deep learning models trained on diverse data that can be easily adapted to a wide range of downstream tasks. An intriguing aspect of these foundation models is their versatility in working with different modalities and tasks. By leveraging the fusion of multiple modalities, these models extract rich representations, enabling tasks such as zero-shot classification. Notable examples of multi-modal foundation models include CLIP~\cite{radford2021learning}, ALIGN~\cite{jia2021scaling}, BASIC~\cite{pham2021combined}, and LiT ~\cite{zhai2022lit}. 
While these models have demonstrated impressive performance across various benchmarks, an important question arises regarding their robustness when faced with uncertainties encountered in real-world applications.

\andrei{Deep Neural Networks} often exhibit limitations in terms of robustness \cite{kendall2017uncertainties, gawlikowski2023survey,franchi2021robust,corbiere2022robust}, necessitating a closer examination of their reliability. Ensuring robustness is essential, as it allows models to maintain reliable performance even when exposed to naturally-induced image corruptions or data distributions that differ from the training distribution. One significant issue pertains to the lack of proper calibration \cite{guo2017calibration}, where the confidence scores of the models should align with their accuracy. \gianni{In this context, our intention is to investigate whether this constraint also affects multi-modal Foundation models, with a particular focus on CLIP. To the best of our knowledge, 
\andrei{this work is among the first to delve into the robustness aspects of CLIP in unsupervised learning}. However, it is worth noting that robustness assumes a pivotal role in real-world scenarios, where multi-modal models are required to adeptly manage a range of uncertainties stemming from data noise, out-of-distribution samples, and adversarial attacks. The capacity to comprehend and quantify these uncertainties is of paramount importance in ensuring robust decision-making and the successful deployment of multi-modal models in critical applications.
}

\begin{figure*}[ht!]
\centering
\begin{tabular}{lll}
\multicolumn{1}{c}{  CLIP }  & 
\multicolumn{1}{c}{ Supervised CLIP + Linear Probing } & 
\multicolumn{1}{c}{  LP-CLIP }  \\
\includegraphics[width=0.3\linewidth]{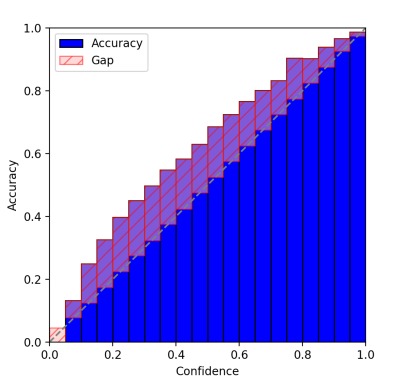} & 
\includegraphics[width=0.3\linewidth]{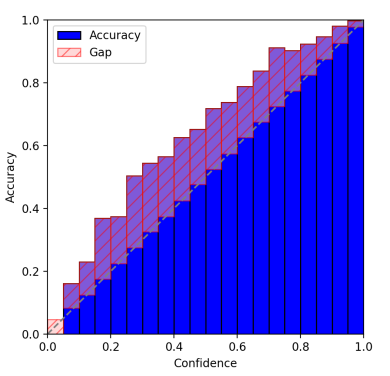} & 
\includegraphics[width=0.3\linewidth]{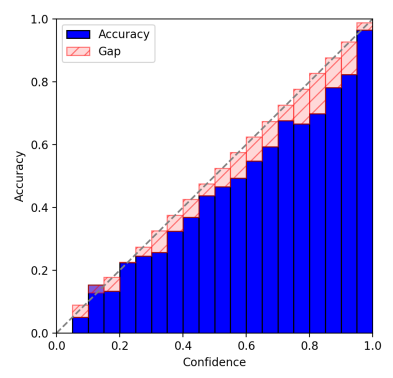} 
\end{tabular}
\caption{
\andrei{\textbf{Reliability diagrams (calibration plots) on CIFAR-10.} We compared calibration of CLIP predictions in zero-shot classification (\emph{left}), supervised linear probing on CLIP features (\emph{middle}) and our LP-CLIP trained in an unsupervised manner on CLIP features (\emph{right}). Overall, LP-CLIP exhibits better calibration properties, while achieving better accuracy for in-domain, domain shift and out-of-distribution settings (see results in section \S\ref{sec:results}).    } 
}
\label{fig:CLIP_Calibration}
\end{figure*}

In light of these challenges, our paper addresses the need for enhancing the robustness of multi-modal foundation models, with a particular focus on CLIP\cite{radford2021learning}. CLIP has garnered attention for its exceptional performance, achieving state-of-the-art zero shot classification results with 64.18\% top-1 accuracy on the ImageNet dataset. Our proposed technique aims to improve the robustness of CLIP through a novel training approach. Specifically, we \gianni{train} %
a linear layer on top of CLIP using an extensive collection of unlabeled images, thereby incorporating unsupervised learning. By leveraging this training set, we can enhance the model's capability to handle uncertainties and improve its robustness in real-world scenarios. The core contribution of our work lies in developing a methodology that optimizes the training procedure of multi-modal foundation models, specifically targeting CLIP. Through our approach, we aim to bridge the gap between the current limitations of multi-modal test-image models and the robustness required for real-world applications. By effectively capturing uncertainties and improving calibration, our technique strives to enhance the reliability and performance of multi-modal models in challenging scenarios.

\textbf{Contributions.} Our work presents several key contributions. Firstly, we are the first to extensively investigate the robustness of CLIP, shedding light on its limitations and potential areas for improvement. Secondly, we propose a novel strategy to enhance the robustness of CLIP, offering practical insights into addressing its vulnerabilities. Lastly, we demonstrate competitive or state-of-the-art results on various datasets, particularly in the accuracy of unsupervised classification, and  in effectiveness of detecting out-of-distribution samples.

\section{Related work}

\paragraph{Uncertainty quantification and DNNs.}
Bayesian Neural Networks (BNNs)\cite{mackay1992practical, neal1995bayesian} have been a fundamental source of inspiration for uncertainty quantification in deep learning. While variational inference\cite{jordan1999introduction, blundell2015weight} has made progress in scaling BNNs, training large DNN architectures with BNNs remains challenging~\cite{dusenberry2020efficient}. Deep Ensembles (DE)\cite{DeepEnsembles} emerged as a practical and efficient instance of BNNs.  MC dropout~\cite{gal2016dropout} is another approach that can be considered to approximate BNNs. Most of these techniques, including ensemble methods, utilize multiple forward passes to quantify the uncertainty of DNNs. Another family of techniques are based on learning the error that the DNN performs~\cite{kendall2017uncertainties,franchi2022latent}, yet these techniques mostly model aleatoric uncertainty. Beyond these, there are approaches that aim to model the limits of knowledge of the DNN. For instance, in ~\cite{franchi2022one,padhy2020revisiting},  One vs ALL training enables modeling the lack of knowledge of a DNN. Sensoy et al.~\cite{sensoy2018evidential} utilized evidential deep learning to model the unknowns of DNNs.

\paragraph{Uncertainty quantification and Foundation Models.}

While foundation models have been extensively studied, there is limited research on their uncertainty quantification. Lee's work~\cite{lee2023mathematical} explores the robustness of GPT, and Pelrine et al.\cite{pelrine2023towards} compare the robustness of various foundation models for NLP tasks. Hendrycks et al.\cite{hendrycks2020pretrained} investigate out-of-distribution (OOD) generalization and OOD detection performance of BERT for multiple NLP tasks. Fort et al.\cite{fort2021exploring} analyze the reliability of OOD detection in ViT models and demonstrate that fine-tuned ViT pre-trained models significantly enhance near OOD detection tasks. Shu et al.\cite{shu2023clipood} propose retraining CLIP with beta moving average to improve its generalization power to domain shift. Esmaeilpour et al.\cite{esmaeilpour2022zero} use a pretrained BERT model to generate OOD classes to quantify the uncertainty of CLIP. Allingham et al.\cite{allingham2023simple} propose enhancing CLIP ensembling to improve OOD detection. However, most of these works focus solely on OOD uncertainty, whereas our research addresses different types of uncertainty.

\begin{figure*}[t!h]
    \centering
\includegraphics[width=0.68\linewidth]{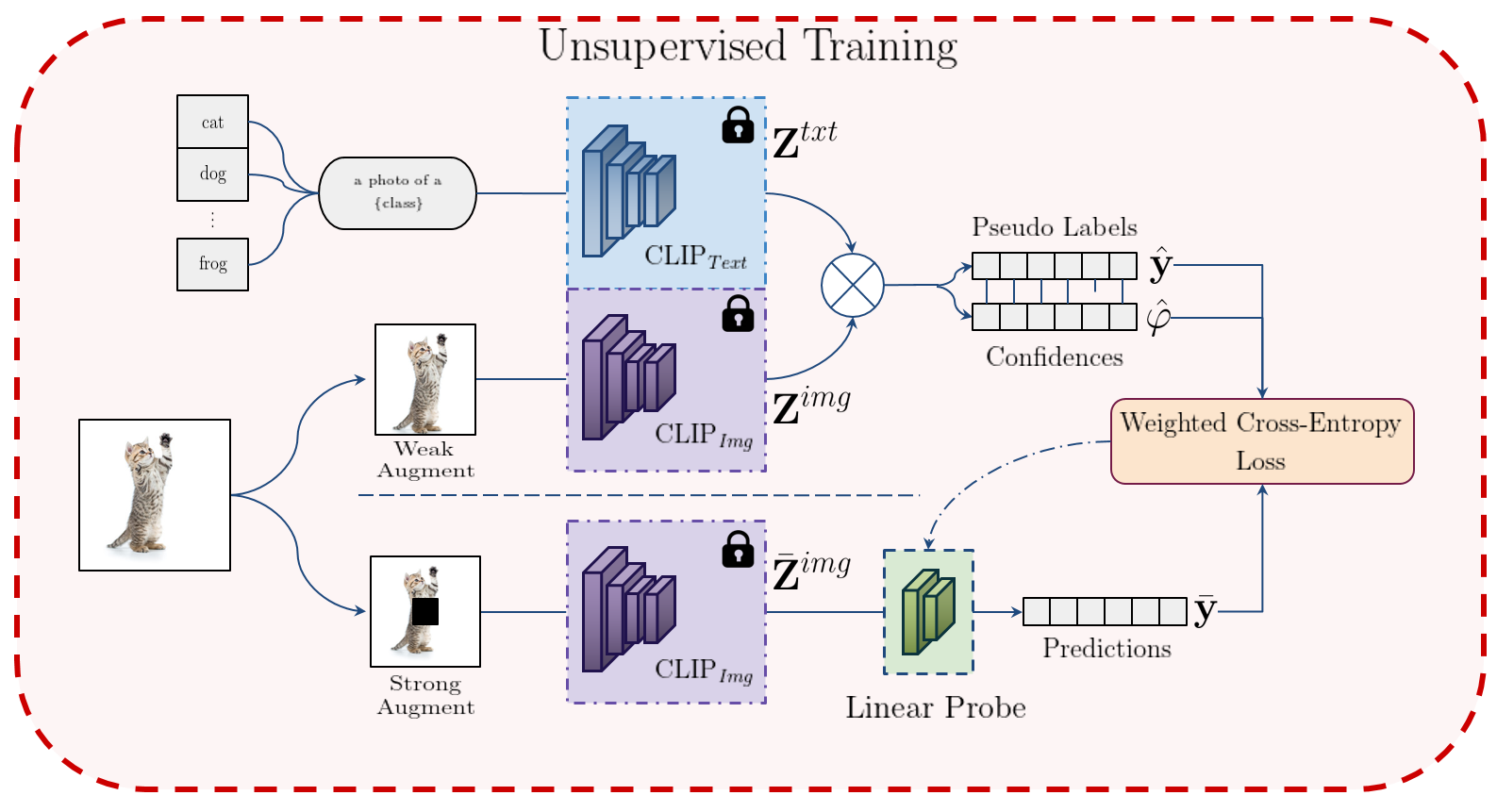}\\
\includegraphics[width=0.68\linewidth]{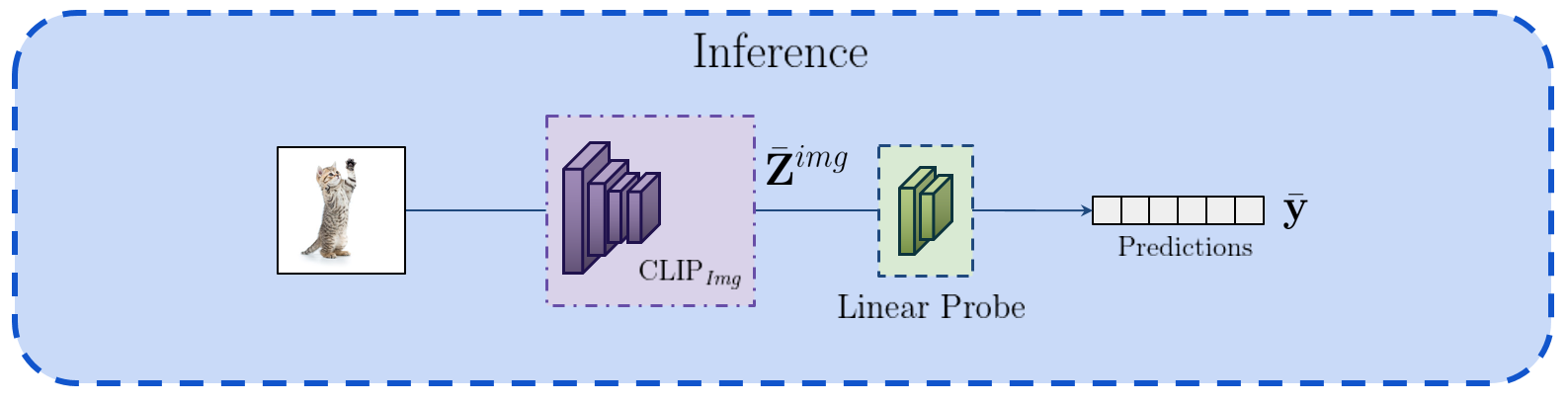}
    \caption{
    \andrei{\textbf{Overview of LP-CLIP, an unsupervised fine-tuning strategy for CLIP.} Given a target dataset without annotations, we generate text embeddings for the classes to recognize and train a linear probe on top of CLIP's image encoder. The linear probe is trained in an unsupervised manner in a teacher-student setting. Our method involves two optimization steps: (1) extracting pseudo-labels using CLIP zero-shot classification and (2) employing the pseudo-labels to train LP-CLIP. At inference, LP-CLIP predicts over the set of classes it was trained upon in the previous stage.}
    }
    \label{fig:diagram}
\end{figure*}

\section{Background}
\subsection{CLIP background and notations}\label{sec:background}
CLIP (Contrastive Language-Image Pretraining) \cite{radford2021learning} is a multi-modal model developed by OpenAI, specifically designed to capture semantic relationships between images and text. By utilizing contrastive pretraining, CLIP learns to align the semantic spaces of text and image through a large corpus of image-text pairs. This pretraining process enables the emergence of strong zero-shot capabilities.

Denoting $\{\vx_i, \vt_i \}_{i=1}^N$ as a set of $N$ image-text pairs, CLIP consists of two separate Deep Neural Networks (DNNs): one for image processing, typically based on a Vision Transformer (ViT) \cite{dosovitskiy2020image}, denoted as $f_{\omega^{img}}^{img}(\cdot)$, and another for text processing, often based on GPT2 \cite{radford2019language}, denoted as $f_{\omega^{txt}}^{txt}(\cdot)$. 
Here, $\omega^{txt}$ and $\omega^{img}$ represent the weights of the text and image DNNs, respectively. By feeding an image $\vx_i$ through the image DNN, we obtain a representation $\vz_i^{img} =f_{\omega^{img}}^{img}(\vx_i)$, and likewise, for text, $\vz_i^{txt} =f_{\omega^{txt}}^{txt}(\vt_i)$. During training, CLIP employs the InfoNCE loss \cite{oord2018representation} to optimize the representations such that $\vz_i^{img}$ and $\vz_i^{txt}$ are close, while $\vz_i^{img}$ and $\vz_j^{txt}$ are distant for $i \neq j$.

The CLIP model has demonstrated impressive generalization properties across various computer vision tasks, including image generation, object detection, and zero-shot image classification \cite{agarwal2021evaluating}. For zero-shot image classification, one needs to use CLIP with 
a set of texts called prompts, such as 
\andrei{``\texttt{This is an image of \{class name\}}'', with the names of the classes used for the classification task. }
Given an image representation $\vz^{img}$ and a set of prompt representations $\{ \vz_c^{txt}\}_{c=1}^C$, where $c$ corresponds to the class ID and $C$ is the total number of classes, we can compute the following logits vector $\vell$:

\begin{equation}
\vell = \left [ \vz^{img} \cdot  \vz_1^{txt}  \ldots  \vz^{img} \cdot \vz_C^{txt} \right ]
\end{equation}

To determine the class, we simply select $\hat{y}=\argmax_c(\vell)$. This process enables zero-shot classification using the learned representations from both the image and text modalities within the CLIP model.

\subsection{Uncertainty background}

Uncertainty in deep learning can primarily stem from three factors, as outlined in a recent survey \cite{gawlikowski2023survey}. Firstly, uncertainty may arise from the data acquisition process. During the acquisition of the training dataset, noise can be introduced due to various real-world factors. For instance, recording training data under specific weather conditions that subsequently change during inference can introduce variability and lead to uncertainty. Additionally, measurement systems themselves, such as sensors, can introduce errors in the acquired data, contributing to uncertainty.

Secondly, uncertainty can arise from the process of building and training deep neural networks (DNNs). DNNs are random functions whose parameters, denoted as $\omega$, are initialized randomly, and the training process relies on stochastic optimization. Consequently, the resulting neural network is a random function that often corresponds to a local minimum of the expected loss function (referred to as the risk). This inherent randomness in the training procedure can introduce errors and uncertainty into the DNN.

The third factor contributing to uncertainty is related to the predictions made by the DNN. Uncertainty can arise from the lack of knowledge within the DNN and may be caused by unknown test data. To categorize the  predictive uncertainty, it is common to separate it into two types: uncertainty caused by the model (referred to as epistemic or model uncertainty) and uncertainty caused by the data (referred to as aleatoric or data uncertainty). Aleatoric uncertainty can be assessed by evaluating the model's performance under various corruptions or noise. However, evaluating robustness under epistemic uncertainty is more challenging, and often the evaluation is limited to assessing the model's performance against out-of-distribution samples.

\begin{table*}[t!]

\begin{center}
\resizebox{0.75\linewidth}{!}{
\begin{tabular}{cccc|rrrr}
\toprule
\multirow{2}{*}{ECE $\downarrow$} & \multirow{2}{*}{Models}             & \multirow{2}{*}{Pretraining datasets} & \multirow{2}{*}{Methods} &\multicolumn{4}{c}{Datasets}  \\
\multicolumn{2}{c}{}                                      &                                       &                          & CIFAR 10 & CIFAR100 & STL10  & TinyImageNet \\
\midrule
\multirow{2}{*}{Supervised}   & \multirow{2}{*}{VIT-B/32} & Imagenet 21K                          & \multirow{2}{*}{LP}      & 0.0418 & 0.2966 & 0.0515 & 0.0764     \\
                              &                           & CLIP                                  &                          & 0.0159 & 0.1131 & 0.0277 & 0.1051       \\
                              \midrule
\multirow{3}{*}{Unsupervised} & \multirow{3}{*}{VIT-B/32} & \multirow{3}{*}{CLIP}                 & Best Prompt              & 0.0587 & 0.1195 & 0.0570 & 0.0637         \\
                              &                           &                                       & Prompt Ensemble          & 0.0587   & 0.1088   & 0.0183 & 0.0192         \\
                              &                           &                                       & LP-CLIP (ours)           & \textbf{0.0203} & \textbf{0.0822} & \textbf{0.0041} & \textbf{0.0174}    \\
                              \midrule
                              \midrule
\multirow{2}{*}{Supervised}   & \clement{VIT-L/16} & Imagenet 21K                          & \multirow{2}{*}{LP}      & 0.0114 & 0.0185 & 0.0169 & 0.0232     \\
                              & \clement{VIT-L/14@336px} & CLIP                                  &                          & 0.0055 & 0.0343 & 0.0113 & 0.0339        \\
                              \midrule
\multirow{3}{*}{Unsupervised} & \multirow{3}{*}{\clement{VIT-L/14@336px}} & \multirow{3}{*}{CLIP}                 & Best Prompt              &0.0574 & 0.0790 & 0.0328 & 0.0378         \\
                              &                           &                                       & Prompt Ensemble          & 0.0427 & 0.0952 & 0.0254 & \textbf{0.0155}        \\
                              &                           &                                       & LP-CLIP (ours)           & \textbf{0.0095} & \textbf{0.0140} & \textbf{0.0080} & 0.0211  \\
                              \bottomrule
\end{tabular}
}

\caption{\textbf{\clement{Expected Calibration Error (ECE)} on the different (in-distribution) datasets}: CIFAR10, CIFAR10, and TinyImageNet, using \clement{ViT-B/32, ViT-L/14 or ViT-L/16} as the backbone. All the results involving training are conducted over three different seeds, and their performances are averaged.}\label{table:tnb1_ECE1}
\end{center}

\end{table*}

\section{Method}

\begin{table*}[t!]
\begin{center}
\resizebox{0.99\linewidth}{!}{
\begin{tabular}{cccc|rrrrrr}
\toprule
\multirow{2}{*}{Accu $\uparrow$} & \multirow{2}{*}{Models}   & \multirow{2}{*}{Pretraining datasets} & \multirow{2}{*}{Methods} & \multicolumn{6}{c}{Datasets}                                                                                                                                                                        \\
                              &                           &                                       &                          & \multicolumn{1}{c}{CIFAR10} & \multicolumn{1}{c}{CIFAR10-C} & \multicolumn{1}{c}{CIFAR100} & \multicolumn{1}{c}{CIFAR100-C} & \multicolumn{1}{c}{TinyImageNet} & \multicolumn{1}{c}{TinyImageNet-C} \\
\midrule
\multirow{2}{*}{Supervised}   & \multirow{2}{*}{VIT-B/32} & Imagenet 21K                          & \multirow{2}{*}{LP}      & 0.9597                      & 0.8266                        & 0.8222                       & 0.6283                        & 0.8312                           & 0.5337                             \\
                              &                           & CLIP                                  &                          & 0.9432                      & 0.7783                        & 0.7649                       & 0.5472                        & 0.7272                           & 0.4427                             \\
\midrule
\multirow{6}{*}{Unsupervised} & \multirow{6}{*}{VIT-B/32} & \multirow{6}{*}{CLIP}                 & Best Prompt   & 0.9001                      & 0.7347                        & 0.6452                       & 0.4490                         & 0.6246                           & 0.3923                             \\
                              &                           &                                       & Prompt Ensemble          & 0.8986                      & 0.7375                        & 0.6507                       & 0.4571                         & 0.6224                           & 0.3973                             \\
                              &                           &                                       & LP w/o WL Aug            & 0.9141                      & 0.7424                        & 0.6569                       & 0.4554                         & 0.6257                           & 0.3861                             \\
                              &                           &                                       & LP w/o Aug               & 0.9097                      & 0.7407                        & 0.6653                       & 0.4643                         & 0.6267                           & 0,3867                            \\
                              &                           &                                       & LP w/o WL                & 0.9257                      & \textbf{0.7768}                        & 0.6619                       & 0.4861                         & \textbf{0.6417}                           & 0.4078                             \\
                              &                           &                                       & LP-CLIP (ours)           & \textbf{0.9254}                      & \textbf{0.7767}                        & \textbf{0.6950}                      & \textbf{0.5191}                         & \textbf{0.6412}                           & \textbf{0.4089} \\
\bottomrule
                              
\end{tabular}
}

\caption{\textbf{Accuracy (Accu) on in-distribution and corrupted datasets}: CIFAR10/CIFAR10-C, CIFAR100/CIFAR100-C, and TinyImageNet/TinyImageNet-C, using ViT-B/32 as the backbone. All the results involving training are conducted over three different seeds, and their performances are averaged.}
\label{table:tnb1_accu_corrupt1}
\end{center}

\end{table*}

In this paper, we introduce a novel unsupervised training approach \gianni{named LP-CLIP}, %
which leverages consistency learning within a teacher-student framework \cite{tarvainen2017mean}. This combination forms an optimization component that serves as a consistency constraint in our DNN training system for image classification. The complete process is \andrei{outlined in Figure~\ref{fig:diagram} and in Algorithm~\ref{algo:lp-clip}}.

\begin{algorithm}[t]
\caption{Consistency loss in LP-CLIP}
\label{algo:lp-clip}
    \small
    \SetAlgoLined
	\SetKwInOut{Input}{Input} 
	\SetKwInOut{Output}{Output} 
	\SetKwInOut{Parameter}{Param}
    \SetKwFunction{StrongAug}{StrongAugment}
	\SetKwFunction{WeakAug}{WeakAugment}
    \SetKwFunction{softmax}{softmax}
	\SetKwFunction{max}{max}
	\SetKwFunction{argmax}{argmax}
	
	\Input{Image $\vx$, class embeddings $\vZ^{txt}$ computed from $C$ class names, structure of $f^{\img}$ and  $g {=} h \circ f^{img}$}
    $\vz^{\img} = f^{\img}(\WeakAug(\vx))$ \tcp{teacher CLIP features}
    $\bar{\vz}^{\img} = f^{\img}(\StrongAug(\vx))$ \tcp{student CLIP features}
    $\vell = \vz^{\img} \cdot (\vZ^{\txt})^{\top}$ \tcp{teacher logits} 
    $\bar{\vell} = h(\bar{\vz}^{\img})$ \tcp{output of linear probe}
    $\hat{y} = \argmax_c(\softmax(\vell))$ \tcp{teacher predicted class}
    $\hat{\varphi} = \max(\softmax(\vell))$ \tcp{teacher confidence of predicted class}
    $\mathcal{L}_{\text{cons}} = - \hat{\varphi} \cdot \hat{y}\cdot\texttt{logsoftmax}(\bar{\vell})$ \tcp{weighted cross-entropy loss}
    \Output{$\mathcal{L}_{\text{cons}}$}

\end{algorithm}

Specifically, our training process involves two simultaneous optimization steps: \textbf{(1)} extracting pseudo-labels using CLIP zero-shot classification and \textbf{(2)} employing teacher-student optimization with strong data augmentation on images provided to the student DNN.

\andrei{In detail,} the teacher DNN is comprised of the CLIP zero-shot classification model explained in section \ref{sec:background}, while the student DNN is denoted as $g_{\omega}(\cdot)= h_{\omega^h} \circ f_{\omega^{img}}^{img}(\cdot)$, where $h_{\omega^h}$ represents a fully connected layer. In our framework, we fix the weights of $f_{\omega^{img}}^{img}$ and solely train the weights of $h_{\omega^h}$. The optimization of $h_{\omega^h}$ is guided by a consistency loss, denoted as $\mathcal{L}_{\text{cons}}$, which involves a weighted cross entropy. The weights are determined by the reliability of the corresponding pseudo label, with the CLIP zero-shot classification confidence score serving as the weight. This confidence score, denoted as $\hat{\varphi}$, is determined by evaluating the maximum value among the softmax obtained from the CLIP zero-shot classification model for a given sample. The significance of using the CLIP confidence score lies in its ability to shed light on CLIP's uncertainty during its pseudo-annotation process. This information is vital for enhancing the robustness of our new model, as it allows us to gain insights into the areas where CLIP may exhibit uncertainty or potential weaknesses. Access to the unreliability of CLIP's predictions enables us to build a more robust model that takes into account the uncertainties inherent in the underlying CLIP model and improves its performance in challenging real-world scenarios. 

In order to improve the generalization capabilities of our DNN, we apply strong data augmentation exclusively to the images presented to the student DNN. However, it is important to note that we only employ weak data augmentation for the images provided to the teacher DNN.
Drawing inspiration from previous works \cite{arazo2020pseudo,sohn2020fixmatch,pham2021meta}, our strong data augmentation encompasses a combination of random augmentation techniques \cite{cubuk2020randaugment}, cutout \cite{devries2017improved}, and the utilization of 
\andrei{common} data augmentation. 
On the other hand, the weak data augmentation is specifically based on the CLIP transformation. By incorporating both weak and strong data augmentations, we aim to strike a balance that prevents overfitting while encouraging the distillation of CLIP's knowledge on augmented data. This mixture of data augmentations plays a crucial role in enhancing the DNN's ability to generalize effectively. By exposing the student DNN to strong data augmentations, it becomes more resilient to variations and uncertainties encountered in real-world data. Conversely, the teacher DNN is presented with less augmented data to retain a clearer representation of the original information. As a result, this complementary use of weak and strong data augmentations not only aids in preventing overfitting but also ensures that CLIP effectively distills its knowledge and adapts to data variations that may arise in practical scenarios.

In summary, our proposed approach involves unsupervised training for CLIP, incorporating consistency learning within a teacher-student framework. This two-step optimization process, along with the utilization of strong data augmentation and the CLIP zero-shot classification confidence scores, contributes to enhancing the performance and generalization of the DNN.

\section{Experiments}

\subsection{\andrei{Experimental setup}}

\smallskip\noindent\textbf{Datasets.~} To evaluate the robustness of CLIP, we conduct experiments on multiple datasets, including CIFAR10 and CIFAR100 \cite{cifar}, Tiny ImageNet \cite{le2015tiny}, STL-10 \cite{coates2011analysis}, and ImageNet \cite{deng2009imagenet}. Additionally, we considered several out-of-distribution (OOD) datasets, namely SVHN \cite{netzer2011reading}, Texture \cite{wang2022vim}, and ImageNet-O \cite{hendrycks2021natural}.

\smallskip\noindent\textbf{Implementation details.~} We use the pre-trained CLIP models, ViT-B/32 and \clement{VIT-L/14@336px}, for our experiments. \clement{CLIP's backbones were pretrained on a 15M subset of YFCC100M \cite{radford2021learning, thomee2016yfcc100m} . We us the first model because it is the most 
\andrei{commonly used} while the latter, which is finetuned on images of size 336x336 pixels, seems to achieve the best performance according to Radford et al. \cite{radford2021learning}. To compare our performance with supervised methods, we look at the same architectures pretrained on ImageNet21K.  To the best of our knowledge, there are no publicly available weights for ViT-L/14 pretrained on ImageNet21K, we therefore used a ViT-L/16 which is the most similar.} 

The temperature parameter of the softmax function remains consistent with the pre-trained models, set at $\tau = 0.01$. 
All models have been trained using the SGD optimizer. We trained all the supervised linear probing with a learning rate of 0.03 and we use a learning rate of 0.1 on the first 3 datasets with the LP-CLIP ViT-B/32. We use a learning rate of 0.01 for other experiences with LP-CLIP. Our technique is implemented in PyTorch, and the complete code will be made available after the anonymity period. 

The performance of CLIP's zero-shot classification heavily relies on the choice of prompt \andrei{and the amount of engineering invested in tuning~\cite{radford2021learning, zhai2022lit, shu2022test, menon2023visual} or learning it~\cite{zhou2022learning,zhou2022conditional}.}
To mitigate the dependence on a single prompt, an effective approach is to utilize an ensemble of prompts. 
\andrei{We do not conduct prompt engineering and to ensure a fair comparison we use } the same prompts as proposed in the original CLIP paper \cite{radford2021learning} for each dataset. In cases where specific prompts are not mentioned, our experiments were conducted with the best-performing prompt \andrei{and} \gianni{we explain how we determine the best prompt in Appendix \textbf{B}}. By incorporating an ensemble of prompts, we aim to enhance the robustness and reliability of CLIP's zero-shot classification across datasets.

\begin{table*}[t!]
\begin{center}
\resizebox{0.90\linewidth}{!}{
\begin{tabular}{cccc|rrrrrr}
\toprule
\multirow{2}{*}{Accu $\uparrow$} & \multirow{2}{*}{Models}   & \multirow{2}{*}{Pretraining datasets} & \multirow{2}{*}{Methods} & \multicolumn{6}{c}{Datasets}                                                                                                                                                                        \\
                              &                           &                                       &                          & \multicolumn{1}{c}{CIFAR10} & \multicolumn{1}{c}{CIFAR10-C} & \multicolumn{1}{c}{CIFAR100} & \multicolumn{1}{c}{CIFAR100-C} & \multicolumn{1}{c}{TinyImageNet} & \multicolumn{1}{c}{TinyImageNet-C} \\
\midrule
\multirow{2}{*}{Supervised}   & \multirow{2}{*}{\clement{VIT-L/14@336px}} & Imagenet 21K                          & \multirow{2}{*}{LP}      & 0.9785                      & 0.8918                        & 0.8862                       & 0.7260                         & 0.8909                           & 0.6174                             \\
                              &                           & CLIP                                  &                          & 0.9759                      & 0.8895                        & 0.8603                       & 0.6880                         & 0.8469                           & 0.6055                             \\
                              \midrule
\multirow{6}{*}{Unsupervised} & \multirow{6}{*}{\clement{VIT-L/14@336px}} & \multirow{6}{*}{CLIP}                 &  Best Prompt   & 0.9468                      & 0.8457                        & 0.7628                       & 0.5972                         & 0.7326                           & 0.5054                             \\
                              &                           &                                       & Prompt Ensemble          & 0.9492                      & 0.8493                        & 0.7703                       & 0.6090                         & 0.7588                           & 0.5326                             \\
                              &                           &                                       & LP w/o WL Aug            & 0.9626                      & 0.8663                        & 0.7916                       & 0.6197                         & 0.7599                           & 0.5229                            \\
                              &                           &                                       & LP w/o Aug               & 0.9681                      & 0.8901                        & 0.7940                       & 0.6448                         & 0.7673                           & 0.5539                             \\
                              &                           &                                       & LP w/o WL                & 0.9623                      & 0.8636                        & 0.7837                       & 0.6087                         & 0.7542                           & 0.5184                             \\
                              &                           &                                       & LP-CLIP (ours)           & \textbf{0.9705}                     & \textbf{0.8980}                        & \textbf{0.8003}                      & \textbf{0.6554}                         & \textbf{0.7761}                           & \textbf{0.5630}   \\
\bottomrule
                              
\end{tabular}
}

\caption{\textbf{Accuracy (Accu) on in-distribution and corrupted datasets}: CIFAR10/CIFAR10-C, CIFAR100/CIFAR100-C, and TinyImageNet/TinyImageNet-C, using ViT-L/14 as the backbone. All the results involving training are conducted over three different seeds, and their performances are averaged.}\label{table:tnb1_accu_corrupt2}
\end{center}
\end{table*}

\begin{table*}[t!]

\begin{center}
\resizebox{0.90\linewidth}{!}{
\begin{tabular}{ccc|rrr|rrr}
\toprule
\multirow{2}{*}{ECE $\downarrow$} & \multirow{2}{*}{Pretraining datasets} & \multirow{2}{*}{Methods} & \multicolumn{3}{c|}{VIT-B/32}                                                                        & \multicolumn{3}{c}{\clement{ViT-L/14@336px -- ViT-L/16}}                                                                        \\
                              &                          &                         & \multicolumn{1}{c}{CIFAR10-C} & \multicolumn{1}{c}{CIFAR100-C} & \multicolumn{1}{c|}{TinyImageNet-C} & \multicolumn{1}{c}{CIFAR10-C} & \multicolumn{1}{c}{CIFAR100-C} & \multicolumn{1}{c}{TinyImageNet-C} \\
\midrule
\multirow{2}{*}{Supervised}   & Imagenet 21K             & \multirow{2}{*}{LP}     & 0.0321                        & 0.0570                         & 0.0770                             & 0.0208                        & 0.0325                         & 0.0668                             \\
                              & CLIP                     &                         & 0.0365                        & 0.1052                         & 0.0984                             & 0.0153                        & 0.0400                         & 0.0847                             \\
\midrule
\multirow{6}{*}{Unsupervised} & \multirow{6}{*}{CLIP}    &   Best Prompt  & 0,0559                        & 0,0711                         & 0.0694                             & 0.0431                        & 0.0530                         & 0,0578                             \\
                              &                          & Prompt Ensemble         & 0.0451                        & 0.0824                         & 0.0510                             & 0.0359                        & 0.0707                         & 0.1665                             \\
                              &                          & LP w/o Aug              & 0.1251                        & 0.1759                         & 0.0757                             & 0.0186                        & 0.0399                         & 0.0639                             \\
                              &                          & LP w/o WL               & \textbf{0.0352}                        & \textbf{0.0323}                         & 0.0669                             & 0.0331                        & \textbf{0.0227}                         & 0.0666                             \\
                              &                          & LP w/o WL Aug           & 0.0552                        & 0.0582                         & 0.1028                             & 0.0413                        & 0.0856                         & 0.0749                             \\
                              &                          & LP-CLIP (ours)          & 0.0739                        & 0.1111                         & \textbf{0.0650}                            & \textbf{0.0121}                        & \textbf{0.0227}                         & \textbf{0.0602}   \\
                              \bottomrule
\end{tabular}
}
\caption{\textbf{Expected Calibration Error (ECE) on corrupted datasets}: CIFAR10/CIFAR10-C, CIFAR100/CIFAR100-C, and TinyImageNet/TinyImageNet-C, using ViT-B/32, ViT-L/14 \clement{or ViT-L/16} as  backbones. All the results that involve training are conducted over three different seeds, and their performances are averaged.}\label{table:tnb1_ECE_corrupt1}
\end{center}

\end{table*}

\smallskip\noindent\textbf{Metrics.~} To assess aleatoric uncertainty, we employ accuracy as the primary criterion. Additionally, we consider the expected calibration error (ECE) \cite{guo2017calibration}, which measures the relationship between the confidence scores predicted by a DNN and its accuracy. We also evaluate cA (accuracy on the corrupted version of the dataset) and cECE (expected calibration error on the corrupted version of the dataset). \clement{The corrupted datasets introduced by Hendrycks et al. \cite{hendrycks2019benchmarking} include various } %
perturbations such as Gaussian noise, shot noise, impulse noise, defocus blur, frosted glass blur, motion blur, zoom blur, snow, frost, fog, brightness, contrast, elastic, pixelate, and JPEG with five different levels of corruption.

Furthermore, to evaluate epistemic uncertainty, we employ the area under the precision-recall curve (AUPR), area under the receiver operating characteristic curve (AUC), and false positive rate at 95\% true positive rate (FPR-95-TPR) as defined in \cite{hendrycks2016baseline}. These metrics provide insights into the DNN's ability to detect OOD data. By analyzing results over multiple metrics, we \andrei{aim for}  a comprehensive assessment of the DNN's performance in terms of accuracy, calibration error, failure rate, and OOD detection capabilities.

\subsection{Main results}
\label{sec:results}

\smallskip\noindent\textbf{In-distribution performance.~} \andrei{We first assess the impact of LP-CLIP in terms of predictive performance and calibration under the same distribution as used for finetuning. Results in Tables~\ref{table:tnb1_ECE1}, 
\ref{table:tnb1_accu_corrupt1}, and \ref{table:tnb1_accu_corrupt2} show consistent improvements over the original CLIP in zero-shot classification with different prompting strategies, even though LP-CLIP uses the same class text embeddings during finetuning of the linear probe. Furthermore, LP-CLIP's performance is not far from the supervised variant that uses ground-truth labels.}

\smallskip\noindent\textbf{Robustness to domain shift.~} To validate the robustness of CLIP under domain shift, we conduct evaluations on variants of the original datasets, namely CIFAR10-C, CIFAR100-C, and TinyImageNet-C. These variants underwent a distribution shift with slight modifications to the images while keeping the labels unchanged. The objective was to observe how CLIP performs when faced with these variations \andrei{away from the original training distribution}.

\begin{figure*}[t]
\centering
\begin{tabular}{lll}
\multicolumn{1}{c}{  CLIP }  & 
\multicolumn{1}{c}{ Supervised CLIP + linear probing } & 
\multicolumn{1}{c}{  LP-CLIP }  \\
\includegraphics[width=0.29\linewidth]{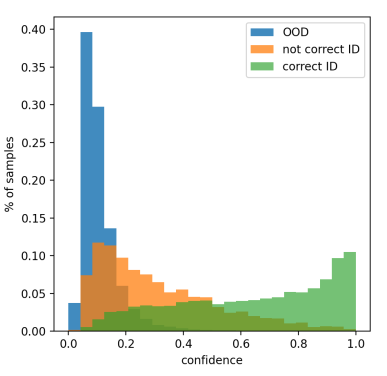} & 
\includegraphics[width=0.29\linewidth]{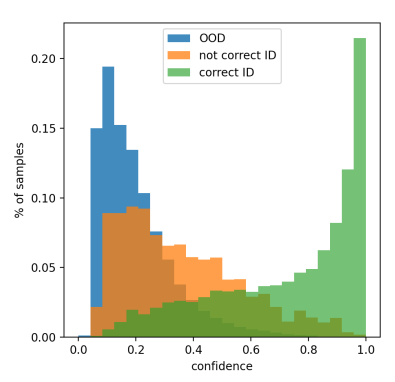} & 
\includegraphics[width=0.29\linewidth]{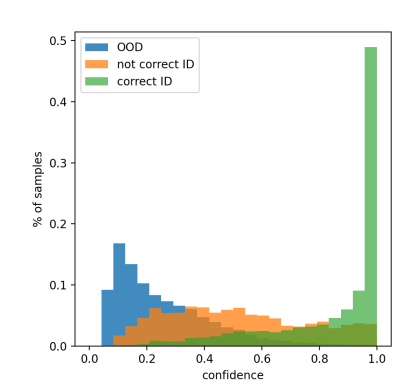} 
\end{tabular}
\caption{\textbf{Confidence histograms on CIFAR-10.} The confidence histograms depict the probabilities of a DNN making predictions with various confidence scores. 
\andrei{The histograms depict (\emph{left}) CLIP zero-shot predictions, (\emph{middle}) supervised linear probing + CLIP, and (\emph{right}) LP-CLIP predictions.}
The histograms are color-coded for clarity. The blue histogram corresponds to the OOD predictions, the orange histogram represents incorrect predictions, and the green histogram indicates correct predictions.
}
\label{fig:CLIP_OOD}
\end{figure*}

Tables \ref{table:tnb1_accu_corrupt1}, \ref{table:tnb1_accu_corrupt2}, and \ref{table:tnb1_ECE_corrupt1} reveal that CLIP experiences accuracy losses of up to 16\% on CIFAR10, 19\% on CIFAR100, and 23\% on TinyImageNet. These accuracy reductions indicate that CLIP struggles to adapt to changes in the training distribution. However, it is noteworthy that CLIP still maintains good 
\andrei{calibration scores} on these corrupted datasets, suggesting that despite the performance decline, CLIP remains fairly reliable.

The variant employing supervised linear probing exhibits similar accuracy losses, around 16\% on CIFAR10, 21\% on CIFAR100, and 28\% on TinyImageNet. In contrast, our approach, LP-CLIP, 
\andrei{shows milder drops in accuracy} of 14\% on CIFAR10, 17\% on CIFAR100, and 23\% on TinyImageNet. This demonstrates that our training technique enhances the generalization capabilities of CLIP. Furthermore, as shown in Tables \ref{table:tnb1_accu_corrupt1} and \ref{table:tnb1_accu_corrupt2}, our performance surpasses that of CLIP without linear probing and comes close to CLIP with total supervision.

Additionally, considering the calibration performance in Table \ref{table:tnb1_ECE_corrupt1}, LP-CLIP exhibits excellent calibration. Therefore, LP-CLIP not only proves to be more robust than CLIP or the fully supervised variant of linear probing but also demonstrates competitive results in terms of robustness and 
\andrei{predictive performance.}

\gianni{Figure \ref{fig:CLIP_Calibration} also indicates that CLIP exhibits relatively good calibration. However, CLIP tends to display overconfidence, wherein it assigns high confidence scores to incorrect predictions, thus making its predictions less reliable. Comparatively, we observe that CLIP demonstrates lower overconfidence levels than its supervised counterpart. This observation is noteworthy as it suggests that CLIP performs reasonably well even without extensive training. On the other hand, LP-CLIP exhibits superior performance by being slightly underconfident. Notably, LP-CLIP achieves better results with a lower calibration error in absolute terms, thus making it a more robust model.}

\smallskip\noindent\textbf{Reliability of epistemic uncertainty.~} \andrei{Here}, we explore the sensitivity of our method in quantifying epistemic uncertainty, specifically focusing on its ability to provide a confidence value for detecting OOD samples.

Figure \ref{fig:CLIP_OOD} presents histograms of confidence bins comparing CLIP predictions with the variant using linear probing and supervised training, and LP-CLIP. The results show that LP-CLIP exhibits higher confidence levels on well-ranked data, making it more reliable than CLIP. Tables \ref{table:OOD1} and \ref{table:OOD2} demonstrate that, in terms of accuracy, LP-CLIP is often comparable to the supervised variant. Notably, when using the ViT-B/32 architecture, CLIP outperforms the supervised variants for TinyImageNet and ImageNet. %
\gianni{It's worth noting that when utilizing ViT-B/32, LP-CLIP and CLIP exhibit equivalence in the worst-case scenario for TinyImageNet and ImageNet datasets.}
However, in most cases, our approach yields results equivalent to the supervised variant. This highlights the usefulness of our technique in detecting anomalous data.

We observed that with VIT-L/14, our approach often outperforms the supervised approach. The difference in performances between ViT-B/32 and VIT-L/14 is intriguing, and we believe this can be attributed to the initial higher accuracy of VIT-L/14. This higher accuracy allows VIT-L/14 to train the linear probing layer more effectively, particularly on challenging datasets. As a result, LP-CLIP benefits from the stronger starting point provided by VIT-L/14, leading to improved performance compared to the supervised approach.

An advantage of our approach over others is that we solely rely on CLIP and do not require additional algorithms such as GPT or others to introduce the notion of OOD. This eliminates potential performance distortions and avoids dependence on other black-box models, the robustness of which would also need further investigation.

\begin{table*}[t!]
\begin{center}
\resizebox{0.90\linewidth}{!}{
\begin{tabular}{cccc|rrrr|rrrr|rrrr}
\toprule
\multicolumn{2}{c}{\multirow{2}{*}{Models}}               & \multirow{2}{*}{Pretraining datasets} & \multirow{2}{*}{Methods}          & \multicolumn{4}{c}{CIFAR10 vs. SVHN}                                                                        & \multicolumn{4}{c}{CIFAR100 vs. SVHN}                                                                       & \multicolumn{4}{c}{STL10 vs. SVHN}                                                                          \\
\multicolumn{2}{c}{}                                      &                                       &                                   & \multicolumn{1}{c}{Accu $\uparrow$} & \multicolumn{1}{c}{ AUC $\uparrow$ } & \multicolumn{1}{c}{AUPR $\uparrow$ } & \multicolumn{1}{c}{FPR95 $\downarrow$} & \multicolumn{1}{c}{Accu $\uparrow$} & \multicolumn{1}{c}{ AUC $\uparrow$ } & \multicolumn{1}{c}{AUPR $\uparrow$ } & \multicolumn{1}{c}{FPR95 $\downarrow$} & \multicolumn{1}{c}{Accu $\uparrow$} & \multicolumn{1}{c}{ AUC $\uparrow$ } & \multicolumn{1}{c}{AUPR $\uparrow$ } & \multicolumn{1}{c}{FPR95 $\downarrow$} \\
\midrule
\multirow{2}{*}{Supervised}   & \multirow{2}{*}{VIT-B/32} & Imagenet 21K                          & \multirow{2}{*}{LP}               & 0,9597                    & 0.9926                  & 0.9873                   & 0.0219                    & 0.8222                    & 0.9575                  & 0.9159                   & 0.2198                    & 0.9903                    & 0.9984                  & 0.9961                   & 0.0030                    \\
                              &                           & CLIP                                  &                                   & 0.9432                    & 0.9890                  & 0.9837                   & 0.0302                    & 0.7649                    & 0.9043                  & 0.8540                   & 0.5565                    & 0.9836                    & 0.9995                  & 0.9991                   & 0.0000                    \\
                              \midrule
\multirow{6}{*}{Unsupervised} & \multirow{6}{*}{VIT-B/32} & \multirow{6}{*}{CLIP}                 & Best Prompt                       & 0.9001                    & 0.9616                  & 0.9473                   & 0.2854                    & 0.6452                    & 0.878                   & 0.8008                   & 0.5879                    & 0.9761                    & 0.9982                  & 0.9961                   & 0.0007                    \\
                              &                           &                                       & Prompt Ensemble                   & 0.8986                    & 0.9843                  & 0.9659                   & 0.1650                    & 0.6507                    & 0.9147                  & 0.8504                   & 0.4361                    & 0.9712                    & 0.9990                  & 0.9977                   & 0.0007                    \\
                              &                           &                                       & LP w/o Aug                        & 0.9097                    & 0.926                   & 0.9152                   & 0.2268                    & 0.6653                    & 0.8463                  & 0.7665                   & 0.6647                    & 0.9797                    & 0.9989                  & 0.9978                   & \textbf{0.0001}                   \\
                              &                           &                                       & LP w/o WL                         & 0.9257                    & 0.9723                  & \textbf{0.9767}                   & 0.2266                    & 0.6619                    & \textbf{0.9227}                  & 0.8636                   & 0.4420                    & 0.9808                    & \textbf{0.9999}                 & \textbf{0.9998}                   & \textbf{0.0001}                    \\
                              &                           &                                       & \multicolumn{1}{l|}{LP w/o WL Aug} & 0.9141                    & 0.9705                  & 0.9744                   & 0.1744                    & 0.6569                    & 0.8623                  & 0.7868                   & 0.6410                    & 0.9797                    & 0.9991                  & 0.9985                   & \textbf{0.0000}                    \\
                              &                           &                                       & LP-CLIP (ours)                    & \textbf{0.9254}                    & \textbf{0.9787}                  & 0.9670                   & \textbf{0.1452}                    & \textbf{0.6950}                    & \textbf{0.9207}                  & \textbf{0.8667}                   & \textbf{0.4163}                    & \textbf{0.9821}                    & \textbf{0.9996}                  & \textbf{0.9990}                   & \textbf{0.0001}                    \\
                              \midrule
                              \midrule
\multirow{2}{*}{Supervised}   & \clement{VIT-L/16} & Imagenet 21K                          & \multirow{2}{*}{LP}               & 0.9785                    & 0.9952                  & 0.9917                   & 0.0066                    & 0.8862                    & 0.9503                  & 0.9025                   & 0.2475                    & 0.9962                    & 0.9990                  & 0.9969                   & 0.0023                    \\
                              &  \clement{VIT-L/14@336px}   & CLIP                                  &                                   & 0.9759                    & 0.9981                  & 0.9970                   & 0.0006                    & 0.8603                    & 0.9772                  & 0.9630                   & 0.1052                    & 0.9971                    & 0.9995                  & 0.9988                   & 0.0000                    \\
                              \midrule
\multirow{6}{*}{Unsupervised} & \multirow{6}{*}{\clement{VIT-L/14@336px}} & \multirow{6}{*}{CLIP}                 & Best Prompt                       & 0.9468                    & 0.9828                  & 0.9698                   & 0.0706                    & 0.7628                    & 0.8443                  & 0.7658                   & 0.7529                    & 0.9939                    & 0.9993                  & 0.9980                   & 0.0021                    \\
                              &                           &                                       & Prompt Ensemble                   & 0.9492                    & 0.9802                  & 0.9665                   & 0.0893                    & 0.7703                    & 0.9092                  & 0.8475                   & 0.4962                    & 0.9944                    & 0.9996                  & 0.9990                   & 0.0001                    \\
                              &                           &                                       & LP w/o Aug                        & 0.9626                    & \textbf{0.9962}                  & \textbf{0.9935}                   & \textbf{0.0082}                    & 0.7916                    & 0.9610                  & \textbf{0.9364}                   & 0.2368                    & \textbf{0.9969}                    & 0.9994                  & 0.9986                   & \textbf{0.0000}                    \\
                              &                           &                                       & LP w/o WL                         & 0.9681                    & \textbf{0.9966}                  & \textbf{0.9930}                   & \textbf{0.0072}                    & 0.7940                    & 0.9625                  & 0.9335                   & 0.2116                    & 0.9967                    & \textbf{0.9998}                  & 0.9994                   & \textbf{0.0000}                    \\
                              &                           &                                       & LP w/o WL Aug                     & 0.9623                    & \textbf{0.9958}                 & \textbf{0.9929}                   & \textbf{0.0029}                    & 0.7837                    & 0.9418                  & 0.9047                   & 0.3466                    & 0.9965                    & 0.9993                  & 0.9983                   & \textbf{0.0000}                    \\
                              &                           &                                       & LP-CLIP (ours)                    & \textbf{0.9705}                    & \textbf{0.9958}                  & \textbf{0.9928}                   & \textbf{0.0081}                    & \textbf{0.8003}                    & \textbf{0.9639}                  & \textbf{0.9344}                   & \textbf{0.1909}                    & \textbf{0.9969}                    & \textbf{0.9998}                 & \textbf{0.9996}                   & \textbf{0.0000}  \\
                              \bottomrule

\end{tabular}
}
\caption{\andrei{\textbf{Comparative results on the OOD detection task.} We evaluate the accuracy, AUC, AUPR and FPR95 on various pairs of in-distribution and OOD datasets CIFAR10 vs. SVHN CIFAR100 vs. SVHN, STL10 vs. SVHN, using ViT-B/32, ViT-L/14 \clement{or ViT-L/16} as backbones. All the results that involve training are conducted over three different seeds, and their performances are averaged.}}\label{table:OOD1}
\end{center}

\end{table*}

\begin{table*}[t!]

\begin{center}
\resizebox{0.90\linewidth}{!}{

\begin{tabular}{cccc|rrrr|rrrr|rrrr}
\toprule
\multicolumn{2}{c}{\multirow{2}{*}{Models}}               & \multirow{2}{*}{Pretraining datasets} & \multirow{2}{*}{Methods}          & \multicolumn{4}{c}{TinyImageNet vs. Texture}                                                                & \multicolumn{4}{c}{ImageNet vs. Texture}                                                                    & \multicolumn{4}{c}{ImageNet vs. ImageNet-O}                                                                 \\

\multicolumn{2}{c}{}                                      &                                       &                                   & \multicolumn{1}{c}{Accu $\uparrow$} & \multicolumn{1}{c}{ AUC $\uparrow$ } & \multicolumn{1}{c}{AUPR $\uparrow$ } & \multicolumn{1}{c}{FPR95 $\downarrow$} & \multicolumn{1}{c}{Accu $\uparrow$} & \multicolumn{1}{c}{ AUC $\uparrow$ } & \multicolumn{1}{c}{AUPR $\uparrow$ } & \multicolumn{1}{c}{FPR95 $\downarrow$} & \multicolumn{1}{c}{Accu $\uparrow$} & \multicolumn{1}{c}{ AUC $\uparrow$ } & \multicolumn{1}{c}{AUPR $\uparrow$ } & \multicolumn{1}{c}{FPR95 $\downarrow$} \\
\midrule
\multirow{2}{*}{Supervised}   & \multirow{2}{*}{VIT-B/32} & Imagenet 21K                          & \multirow{2}{*}{LP}               & 0,8312                    & 0.9557                  & 0.9907                   & 0.2122                    & 0.7300                    & 0.8612                  & 0.9929                   & 0.5514                    & 0.7300                    & 0.8735                  & 0.9932                   & 0.4837                    \\
                              &                           & CLIP                                  &                                   & 0.7272                    & 0.7654                  & 0.9460                   & 0.8585                    & 0.6897                    & 0.7669                  & 0.9862                   & 0.7697                    & 0.6897                    & 0.6925                  & 0.9819                   & 0.8863                    \\
                              \midrule
\multirow{6}{*}{Unsupervised} & \multirow{6}{*}{VIT-B/32} & \multirow{6}{*}{CLIP}                 & Best Prompt                       & 0.6246                    & \textbf{0.7473}                  & 0.9357                   & \textbf{0.8378}                    & 0.6280                    & \textbf{0.7612}                  & \textbf{0.9860}                   & \textbf{0.7106}                    & \textbf{0.6280}                    & 0.7027                  & 0.9819                   & \textbf{0.8155}                    \\
                              &                           &                                       & Prompt Ensemble                   & 0.6224                    & 0.7434                  & 0.9374                   & 0.8697                    & 0.6335                    & 0.7585                  & 0.9855                   & 0.6968                    & 0.6335                    & \textbf{0.7050}                  & \textbf{0.9822}                   & 0.8190                    \\
                              &                           &                                       & LP w/o Aug                        & 0.6267                    & 0.7277                  & 0.9334                   & 0.8606                    & 0.6447                    & 0.7282                  & 0.9839                   & 0.7973                    & 0.6447                    & 0.6718                  & 0.9721                   & 0.8662                    \\
                              &                           &                                       & LP w/o WL                         & 0.6417                    & 0.7334                  & 0.9353                   & 0.8541                    & 0.6283                    & 0.7584                 & 0.9844                   & 0.7431                    & 0.6283                    & 0.6878                  & 0.981                    & 0.8615                    \\
                              &                           &                                       & \multicolumn{1}{l|}{LP w/o WL Aug} & 0.6257                    & 0.7119                  & 0.9264                   & 0.8702                    & 0.6291                    & 0.7407                  & 0.9847                   & 0.7871                    & 0.6291                    & 0.6845                  & 0.9809                   & 0.8637                    \\
                              &                           &                                       & LP-CLIP (ours)                    & \textbf{0.6412}                    & \textbf{0.7451}                  & \textbf{0.9400}                   & 0.8548                    & \textbf{0.6536}                    & 0.7351                  & 0.9839                   & 0.7760                    & 0.6536                    & 0.6810                  & 0.9806                   & 0.8570                    \\
                              \midrule
                              \midrule
\multirow{2}{*}{Supervised}   & \clement{VIT-L/16} & Imagenet 21K                          & \multirow{2}{*}{LP}               & 0.8909                    & 0.9563                  & 0.9906                   & 0.1991                    & 0.8093                    & 0.8726                  & 0.9931                   & 0.4668                    & 0.8093                    & 0.9207                  & 0.9957                   & 0.3053                    \\
                              &  \clement{VIT-L/14@336px}  & CLIP                                  &                                   & 0.8469                    & 0.8553                  & 0.9661                   & 0.6917                    & 0.8378                    & 0.8450                  & 0.9910                   & 0.5200                    & 0.8378                    & 0.8342                  & 0.9910                   & 0.6058                    \\
                              \midrule
                              
\multirow{6}{*}{Unsupervised} & \multirow{6}{*}{\clement{VIT-L/14@336px}} & \multirow{6}{*}{CLIP}                 & Best Prompt                       & 0.7326                    & 0.7896                  & 0.9486                   & 0.7830                    & 0.7543                    & 0.8086                  & 0.9890                   & 0.6074                    & 0.7543                    & 0.8103                  & 0.9893                   & 0.6545                    \\
                              &                           &                                       & Prompt Ensemble                   & 0.7588                    & 0.7936                  & 0.9478                   & 0.7713                    & 0.7656                    & 0.8033                  & 0.9888                   & 0.6133                    & 0.7656                    & 0.8026                  & 0.9889                   & 0.6490                    \\
                              &                           &                                       & LP w/o Aug                        & 0.7599                    & 0.8252                  & \textbf{0.9622}                   & 0.7160                    & 0.7750                    & 0.8143                  & 0.9890                   & 0.5872                    & 0.7750                    & 0.8184                  & 0.9901                   & 0.6555                    \\
                              &                           &                                       & LP w/o WL                         & 0.7673                    & 0.813                   & 0.9575                   & 0.7388                    & 0.7767                    & 0.8128                  & 0.9890                   & 0.5628                    & 0.7767                    & 0.8121                  & \textbf{0.9906}                   & \textbf{0.6320}                    \\
                              &                           &                                       & LP w/o WL Aug                     & 0.7542                    & 0.8110                  & 0.9558                   & 0.7340                    & 0.7727                    & \textbf{0.8139}                  & \textbf{0.9895}                   & \textbf{0.5649}                    & 0.7727                    & 0.8136                  & \textbf{0.9904}                   & \textbf{0.6320}                    \\
                              &                           &                                       & LP-CLIP (ours)                    & \textbf{0.7761}                    & \textbf{0.8292}                  & \textbf{0.9615}                   & \textbf{0.7113}                    & \textbf{0.7799}                    & \textbf{0.8131}                  & \textbf{0.9895}                   & 0.6044                    & \textbf{0.7799}                    & \textbf{0.8199}                  & \textbf{0.9903}                   & 0.6563 \\
                              \bottomrule
\end{tabular}

}
\caption{\andrei{\textbf{Comparative results on the OOD detection task.}  We evaluate the accuracy, AUC, AUPR and FPR95 on various pairs of in-distribution and OOD datasets TinyImageNet vs. Texture ImageNet vs. Texture and ImageNet vs. ImageNet-O, using ViT-B/32, ViT-L/14 \clement{or ViT-L/16} as  backbones. All the results that involve training are conducted over three different seeds, and their performances are averaged.}}\label{table:OOD2}
\end{center}
\end{table*}

\section{Discussions}

\gianni{
Our study demonstrates that LP-CLIP serves as a powerful tool to enhance the robustness of CLIP and to yield a confidence score that is more faithful to the actual quality of its predictions. Notably, LP-CLIP consistently exhibits superior performance in identifying OOD data in most cases (see Tables \ref{table:OOD1} and \ref{table:OOD2}).}

\gianni{In Appendix \textbf{C}, we meticulously demonstrate the robustness of our technique by subjecting it to changes in the CLIP training dataset. The results are strikingly similar to those achieved with OpenAI's CLIP, underscoring that our approach is not confined to just OpenAI's CLIP architecture.}

\gianni{Furthermore, in Appendix \textbf{D}, we delve into the latent space by visualizing the logits of both CLIP and LP-CLIP. Interestingly, the visualization portrays LP-CLIP's latent space as more organized and structured. This observation suggests that LP-CLIP introduces a denoising effect to the latent space, potentially contributing to its improved performance in various tasks.}

\section{Conclusions}

\gianni{This paper delves into the inherent robustness issues present in the CLIP model, revealing its tendency to generate unreliable zero-shot classification predictions. Moreover, we note that the pretraining of CLIP can sometimes be less robust compared to the ImageNet 21K pretraining. Recognizing the versatility of CLIP, which allows for usage without annotations, we endeavor to bolster its robustness.}

\gianni{To this end, we introduce a straightforward yet effective approach aimed at enhancing CLIP's robustness. This technique involves the integration of a linear probing layer, meticulously trained using pseudo annotations generated through a consistency learning mechanism extracted from CLIP. Through extensive experimentation, we substantiate the capabilities of our proposed LP-CLIP technique. Surpassing the performance of zero-shot CLIP, LP-CLIP even occasionally outperforms the supervised linear probing CLIP, all the while obviating the need for ground truth labels. However, it's important to acknowledge that, like many existing state-of-the-art methods, LP-CLIP does possess a limitation: it lacks theoretical guarantees that ensure the precision of predicted uncertainty.}

\gianni{Looking ahead, our research trajectory will be geared towards exploring the potential integration of LP-CLIP with active learning or incremental learning paradigms. Such an integration could significantly amplify its utility, enabling the provision of robust annotations without requiring human intervention. This future direction aligns with our commitment to continually enhance the practical applicability and adaptability of the LP-CLIP framework.}

\textbf{Acknowledgement:}
This work was performed using HPC resources from GENCI-IDRIS (Grant 2021 - AD011011970R1) and (Grant 2022 - AD011011970R2).
\clearpage
{
\small
\bibliographystyle{ieee_fullname}
\bibliography{egbib}
}
\balance

\clearpage

\appendix

\section{Training hyperparameters}
As delineated in the main paper, the training of our LP-CLIP necessitated the utilization of diverse hyperparameters, which are comprehensively documented in Table \ref{table:hyperparameters}. Throughout the training process, we adopted the SGD optimizer with a momentum value of 0.9 and omitted any weight decay. Employing a Cosine Scheduler and capping the gradient norm at 1, we conducted multiple runs while averaging the outcomes across seeds 42, 36, and 12.

In the training of our LP-CLIP model, we applied a set of robust augmentations exclusively to the data directed to the CLIP student. These augmentations were implemented right before the model's preprocessing transformations. Sequentially, we employed the following augmentations:

\begin{itemize}
    \setlength\itemsep{-1.35mm}
    \item[] Resize(img\_size)
    \item[] RandomCrop(img\_size, padding=4)
    \item[] RandomHorizontalFlip()
    \item[] RandAugment()
    \item[] Cutout(patch\_size = img\_size // 8)
\end{itemize}

\noindent where \textit{img\_size} is the desired size for dataset images before preprocessing. These are all transformations implemented in torchvision, except Cutout introduced by DeVries et al. in \cite{devries2017improved}.

\section{Selected prompts}

\begin{table}[t!h]
\begin{center}
\resizebox{0.99\linewidth}{!}{
\begin{tabular}{llllllll}
\toprule
\multicolumn{3}{c}{Dataset} & \crot{CIFAR10} & \crot{CIFAR100} & \crot{STL10} & \crot{TinyImageNet} & \crot{ImageNet} \\
\midrule
\multirow{6}{*}{\crot{LP-CLIP}} & \multirow{3}{*}{\crot{\small ViT-B/32}} & lr & 0.1 & 0.1 & 0.1 & 0.01 & 0.01 \\
& & max-steps & 15 000 & 20 000 & 1 500 & 31 250 & 400 000 \\
& & batch-size & \multicolumn{5}{c}{64} \\
\cline{2-8} 
& \multirow{3}{*}{\crot{\small ViT-L/14}} & lr & \multicolumn{5}{c}{0.01} \\
& & max-steps & 15 000 & 20 000 & 1 500 & 31 250 & 400 000 \\
& & batch-size & \multicolumn{5}{c}{64} \\
\midrule
\multirow{6}{*}{\crot{Supervised}} & \multirow{3}{*}{\crot{\small ViT-B/32}} & lr & \multicolumn{5}{c}{0.03} \\
& & max-steps & 2 000 & 4 000 & 300 & 7 000 & 20 000 \\
& & batch-size & \multicolumn{5}{c}{512} \\
\cline{2-8} 
& \multirow{3}{*}{\crot{\small ViT-L/16}} & lr & \multicolumn{5}{c}{0.03} \\
& & max-steps & 2 000 & 4 000 & 300 & 7 000 & 20 000 \\
& & batch-size & \multicolumn{5}{c}{512} \\
\bottomrule
\end{tabular}
}
\caption{\textbf{Hyperparameters} used to train LP-CLIP ViT-B/32 and ViT-L/14@336px, and the linear probes on ViT-B/32 and ViT-L/16 pretrained on ImageNet21k and by openAI for CLIP.}\label{table:hyperparameters}

\end{center}\end{table}

Initially, we began with a classical prompt, "a photo of a $<\text{class}>$," which demonstrated effectiveness across many datasets. However, recognizing that this may not be the best prompt, we propose an ensemble prompt approach for each dataset (similarly to \cite{radford2021learning}). In this approach, we employ the official prompts provided by OpenAI as part of the official CLIP package\footnote{https://github.com/openai/CLIP}.

To refine our choice of the best prompt, we engaged in prompt engineering. The challenge with the official CLIP prompts lies in the fact that while some datasets have up to 80 prompts available, others might only offer two prompts. This disparity can lead to suboptimal prompt selection. To address this, we expanded our prompt selection strategy. Initially, we incorporated the 80 prompts designated for the ImageNet dataset into all other datasets. Subsequently, we introduced an additional 38 prompts, including 9 hand-designed prompts and 29 generated prompts using ChatGPT (refer to Table \ref{table:additional_prompts}). This comprehensive list now includes a total of 118 prompts.

In the experimental section, when we mention the "best prompt," we are referring to the prompt that was determined as the most effective for each dataset. The process of selecting the best prompt involves utilizing the ViT-B/32 architecture in conjunction with the set of 118 prompts concatenated to the official prompts of the datase from the official CLIP package. Through this process, we identified the prompt that resulted in the highest zero-shot accuracy for each specific dataset. The best prompts for each dataset are outlined in Table \ref{table:best_prompts}.

\begin{table}[t!]
\begin{center}
\resizebox{0.99\linewidth}{!}{
\begin{tabular}{ll}
\toprule
Dataset & Best prompt \\
\midrule
CIFAR10\cite{krizhevsky2009learning} & "a photograph of a $<\text{class}>$"\\
CIFAR100\cite{krizhevsky2009learning} & "a low contrast photo of a $<\text{class}>$." \\
STL-10\cite{coates2011analysis} & "a capture of a $<\text{class}>$" \\ 
Tiny ImageNet\cite{le2015tiny} & "A photo containing a $<\text{class}>$" \\
ImageNet\cite{deng2009imagenet} & "a good photo of a $<\text{class}>$."\\    
\bottomrule
\end{tabular}}
\caption{The best prompt used for each dataset}\label{table:best_prompts}
\end{center}\end{table}

\begin{table}[t!]
\resizebox{0.99\linewidth}{!}{
\begin{tabular}{l}
\toprule
Hand-crafted prompts \\
\midrule
"\{\}" \\
"a \{\}" \\
"a photo of a \{\}" \\
"a photo with a \{\}" \\
"a photo containing a \{\}" \\
"a photo that contains a \{\}" \\
"a photo with multiple \{\}s" \\
"a photo with many \{\}s" \\
"a photo with some \{\}" \\
\midrule
ChatGPT prompts \\
\midrule
"an image of a \{\}" \\
"a picture of a \{\}" \\
"a snapshot of a \{\}" \\
"a photograph of a \{\}" \\
"a portrait of a \{\}" \\
"a still shot of a \{\}" \\
"a capture of a \{\}" \\
"a visual of a \{\}" \\
"a depiction of a \{\}" \\
"a shot of a \{\}" \\
"a capture of a \{\} in action" \\
"a still image of a \{\}" \\
"an illustration of a \{\}" \\
"a snapshot of a \{\} in its natural habitat" \\
"a photograph capturing a \{\}" \\
"a candid photograph of a \{\}" \\
"a close-up of a \{\}" \\
"an image featuring a \{\}" \\
"a picture showcasing a \{\}" \\
"can you classify this picture as a \{\}?" \\
"what do you think is in this photo? Hint: it's a \{\}." \\
"i'm trying to identify the \{\} in this image." \\
"this photograph features a \{\}, can you tell me what it is?" \\
"in this image, you'll see a \{\}." \\
"can you determine the subject of this photo? It's a \{\}." \\
"i need your help identifying the \{\} in this picture." \\
"this snapshot features a \{\} - can you name it?" \\
"this picture showcases a \{\}, can you classify it?" \\
"i'm trying to find out what kind of creature is in this photograph - it's a \{\}." \\
\bottomrule
\end{tabular}
}
\caption{\textbf{Additional Prompts:} The 9 hand-crafted prompts and the 29 prompts generated by ChatGPT used to find the best prompt}\label{table:additional_prompts}
\end{table}

\section{Experiments with different training datasets}

To ensure the broad applicability of our approach, we examined whether our method exhibits similar efficacy on a CLIP model pre-trained on different alternative datasets. To this end, we replicated our experiments using the ViT-B/32 model as proposed by OpenCLIP \cite{cherti2023reproducible}, which had been pre-trained on datasets such as Laion 400M \cite{schuhmann2021laion} and Laion 2B \cite{schuhmann2022laion}. These datasets are comparable in size to those employed by OpenAI.

The results, as presented in Tables \ref{table:ECE_openclip} and \ref{table:OOD_openclip}, show that the method is insensitive to the training dataset, thereby demonstrating the consistency of our findings. Notably, the expected calibration error (ECE) consistently favors our method over using the best prompt for the CLIP model (as indicated in Table \ref{table:ECE_openclip}), showcasing strong calibration performance. Similarly, with respect to the out-of-distribution (OOD) detection task, LP-CLIP demonstrates improved OOD sample detection in the majority of cases, accompanied by a significant increase in accuracy regardless of the training datasets (as reflected in Table \ref{table:ECE_openclip}).

\begin{table}[t!]

\begin{center}
\resizebox{0.99\linewidth}{!}{
\begin{tabular}{cc|ccc|ccc}
\toprule
\multirow{2}{*}{ECE $\downarrow$} & \multirow{2}{*}{Methods} & \multicolumn{3}{c|}{CIFAR100}  & \multicolumn{3}{c}{TinyImageNet} \\
  &  & \multicolumn{1}{c}{OpenAI} & LAION400M & \multicolumn{1}{c|}{LAION2B} & OpenAI & LAION400M & LAION2B\\
\midrule
\multirow{5}{*}{\crot{Supervised}}  
 &&&&&\\
 &&&&&\\
 & \large LP &\large 0,1131 &\large 0,0126 &\large 0,0114 &\large 0,1051 &\large 0,0126 &\large 0,0125\\
 &&&&&\\
 &&&&&\\
\midrule
\multirow{5}{*}{\crot{Unsupervised}}
 &&&&&\\
 & \large Best Prompt &\large 0,1195 &\large 0,0825 &\large 0,0662 &\large 0,0637 &\large 0,1155  &\large 0,0946  \\
 &&&&&\\
 & \large LP-CLIP     &\large\textbf{0,0822} &\large\textbf{0,0810} &\large\textbf{0,0250} &\large\textbf{0,0174}  &\large\textbf{0,0887}  &\large\textbf{0,0341} \\
 &&&&&\\
\bottomrule
\end{tabular}
}
\caption{\textbf{Expected Calibration Error (ECE)}: datasets CIFAR100 and TinyImageNet using ViT-B/32 as backbone pretrained on OpenAI dataset / LAION400M / LAION2B.}\label{table:ECE_openclip}
\end{center}
\end{table}

\begin{table*}[t!]

\begin{center}
\resizebox{0.99\linewidth}{!}{
\begin{tabular}{cc|cccc|cccc|cccc}
\toprule
\multicolumn{2}{c|}{\multirow{3}{*}{Methods}} & \multicolumn{12}{c}{CIFAR100 vs SVHN} \\
&& \multicolumn{4}{c}{OpenAI} & \multicolumn{4}{c}{LAION-400M} &  \multicolumn{4}{c}{LAION-2B} \\
&& Accu $\uparrow$ & AUC $\uparrow$ & AUPR $\uparrow$ & FPR95 $\downarrow$& Accu $\uparrow$ & AUC $\uparrow$ & AUPR $\uparrow$ & FPR95 $\downarrow$ & Accu $\uparrow$ & AUC $\uparrow$ & AUPR $\uparrow$ & FPR95 $\downarrow$  \\
\midrule
\multirow{5}{*}{\crot{Supervised}}  
 &&&&&&&&&&&&&\\
 &&&&&&&&&&&&&\\
 & \large LP &\large 0,7649&\large0,9043&\large0,854&\large0,5565&\large0,8214&\large0,9109&\large0,8710&\large0,5480&\large0,8596&\large0,9754&\large0,9533&\large0,1375\\
 &&&&&&&&&&&&&\\
 &&&&&&&&&&&&&\\
\midrule
\multirow{5}{*}{\crot{Unsupervised}}
 &&&&&&&&&&&&&\\
 & \large Best Prompt &\large0,6452&\large0,8780&\large0,8008&\large0,5879&\large0,6744&\large\textbf{0,9523}&\large\textbf{0,9072}&\large\textbf{0,2517}&\large0,7504&\large\textbf{0,9727}&\large\textbf{0,9443}&\large\textbf{0,1465}\\
 &&&&&&&&&&&&&\\
 & \large LP-CLIP     &\large\textbf{0,6950}&\large\textbf{0,9207}&\large\textbf{0,8667}&\large\textbf{0,4163}&\large\textbf{0,7171}&\large0,9053&\large0,8643&\large0,5481&\large\textbf{0,7891}&\large0,9707&\large0,9409&\large0,1525\\
 &&&&&&&&&&&&&\\
\bottomrule
\toprule
\multicolumn{2}{c|}{\multirow{3}{*}{Methods}} & \multicolumn{12}{c}{TinyImageNet vs Texture} \\
&& \multicolumn{4}{c}{OpenAI} & \multicolumn{4}{c}{LAION-400M} &  \multicolumn{4}{c}{LAION-2B} \\
&& Accu $\uparrow$ & AUC $\uparrow$ & AUPR $\uparrow$ & FPR95 $\downarrow$& Accu $\uparrow$ & AUC $\uparrow$ & AUPR $\uparrow$ & FPR95 $\downarrow$ & Accu $\uparrow$ & AUC $\uparrow$ & AUPR $\uparrow$ & FPR95 $\downarrow$  \\
\midrule
\multirow{5}{*}{\crot{Supervised}}  
 &&&&&&&&&&&&&\\
 &&&&&&&&&&&&&\\
 & \large LP &\large0,7272&\large0,7654&\large0,9460&\large0,8585&\large0,7582&\large0,7803&\large0,9481&\large0,8000&\large0,7860&\large0,8132&\large0,9568&\large0,7622\\
 &&&&&&&&&&&&&\\
 &&&&&&&&&&&&&\\
\midrule
\multirow{5}{*}{\crot{Unsupervised}}
 &&&&&&&&&&&&&\\
 & \large Best Prompt &\large0,6246&\large\textbf{0,7473}&\large0,9357&\large\textbf{0,8378}&\large0,5965&\large0,6941&\large0,9178&\large0,8846&\large0,6745&\large0,7750&\large0,9416&\large\textbf{0,7394}\\
 &&&&&&&&&&&&&\\
 & \large LP-CLIP     &\large\textbf{0,6412}&\large0,7451&\large\textbf{0,9400}&\large0,8548&\large\textbf{0,6484}&\large\textbf{0,7829}&\large\textbf{0,9436}&\large\textbf{0,7851}&\large\textbf{0,7075}&\large\textbf{0,8059}&\large\textbf{0,9538}&\large0,7468\\
 &&&&&&&&&&&&&\\
\bottomrule
\end{tabular}
}
\caption{\textbf{Comparative results on OOD task.} We evaluate the accuracy, AUC, AUPR and FPR95 on datasets CIFAR100 vs SVHN and TinyImageNet vs Texture, using ViT-B/32 as backbone pretrained on OpenAI dataset / LAION400M / LAION2B.}\label{table:OOD_openclip}
\end{center}
\end{table*}

\section{Latent space visualization}

\begin{figure*}[h!t]
    \centering
    \includegraphics[width=0.9\linewidth]{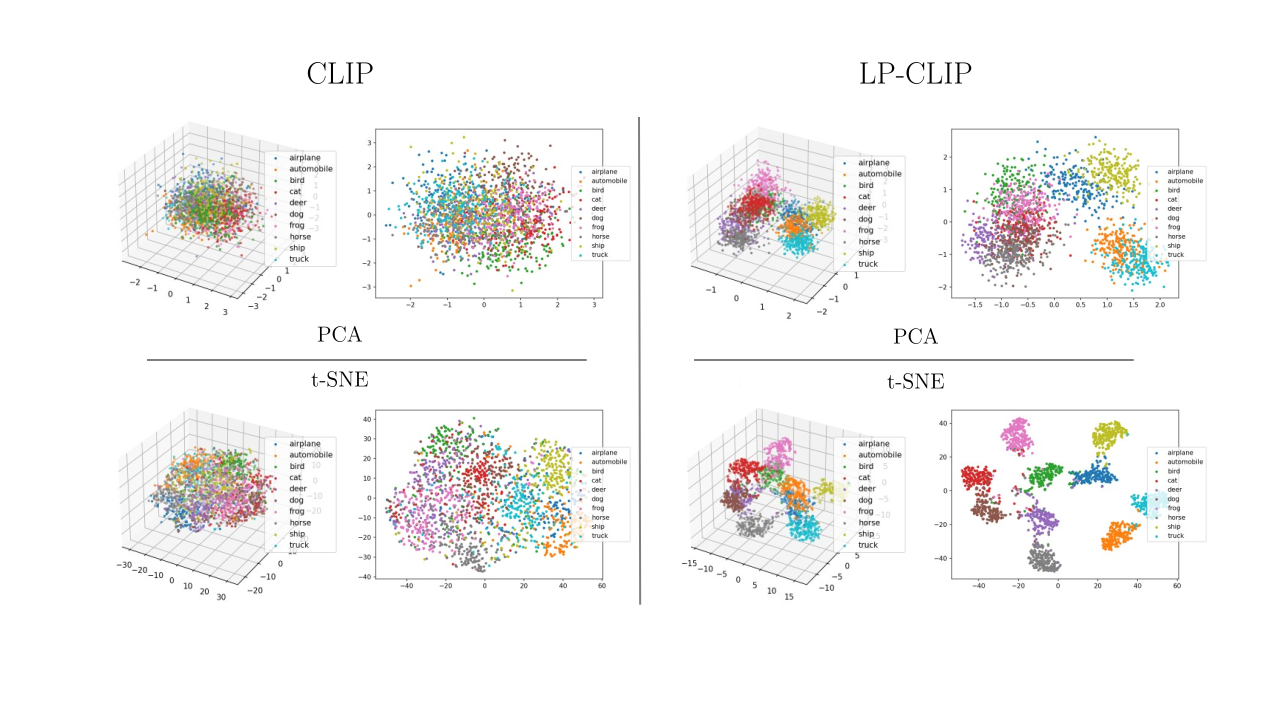}
    \caption{\textbf{Latent Space Visualisation:} PCA and t-SNE reduction of the latent space in 2D and 3D of 2000 sample from the CIFAR10 dataset, for CLIP (left) and LP-CLIP (right) using ViT-B/32 as backbone.}
    \label{fig:latentspace}
\end{figure*}

To gain deeper insights into the representation capabilities of various models, we delved into the latent space. Our analysis involved employing dimensionality reduction techniques, specifically Principal Component Analysis (PCA) and t-Distributed Stochastic Neighbor Embedding (t-SNE), as showcased in Figure \ref{fig:latentspace}. The visualization of these latent spaces provides a clearer perspective on the efficacy of different models.

In particular, the visualizations reveal that the representation generated by our LP-CLIP method exhibits remarkable clustering, indicating that our approach effectively captures the underlying structure of the data. In contrast, achieving similar coherent clusters using the internal representation of the original CLIP model proves challenging. This observation underscores that our LP-CLIP technique excels in distilling CLIP's broad domain knowledge for targeted tasks, all accomplished through an unsupervised framework.

\end{document}